\documentclass[10pt,twocolumn,letterpaper]{article}

\usepackage{cvpr}
\usepackage{times}
\usepackage{epsfig}
\usepackage{graphicx}
\usepackage{amsmath}
\usepackage{amssymb}
\usepackage{caption}
\usepackage{verbatim}
\usepackage{multirow}

\usepackage[breaklinks=true,bookmarks=false]{hyperref}

\cvprfinalcopy 

\def\cvprPaperID{****} 

\ifcvprfinal\fi
\begin{document}

\title{Dynamic Zoom-in Network for Fast Object Detection in Large Images}

\author{Mingfei Gao$^1$~~~Ruichi Yu$^1$~~~Ang Li$^2$\thanks{The work was done while the author was at the University of Maryland}~~~Vlad I. Morariu$^{3*}$~~~Larry S. Davis$^1$\\
$^1$University of Maryland, College Park~~~$^2$DeepMind~~~~$^3$Adobe Research\\
{\tt\small \{mgao,richyu,lsd\}@umiacs.umd.edu~~anglili@google.com~~morariu@adobe.com}
}

\maketitle

\begin{abstract}
   We introduce a generic framework that reduces the computational cost of object detection while retaining accuracy for scenarios where objects with varied sizes appear in high resolution images. Detection progresses in a coarse-to-fine manner, first on a down-sampled version of the image and then on a sequence of higher resolution regions identified as likely to improve the detection accuracy. Built upon reinforcement learning, our approach consists of a model (R-net) that uses coarse detection results to predict the potential accuracy gain for analyzing a region at a higher resolution and another model (Q-net) that sequentially selects regions to zoom in. Experiments on the Caltech Pedestrians dataset show that our approach reduces the number of processed pixels by over $50\%$ without a drop in detection accuracy. The merits of our approach become more significant on a high resolution test set collected from YFCC100M dataset, where our approach maintains high detection performance while reducing the number of processed pixels by about $70\%$ and the detection time by over $50\%$.
\end{abstract}

\section{Introduction}
\label{sec: intro}
\begin{figure}[t]
\begin{center}
   \includegraphics[width=0.9\linewidth]{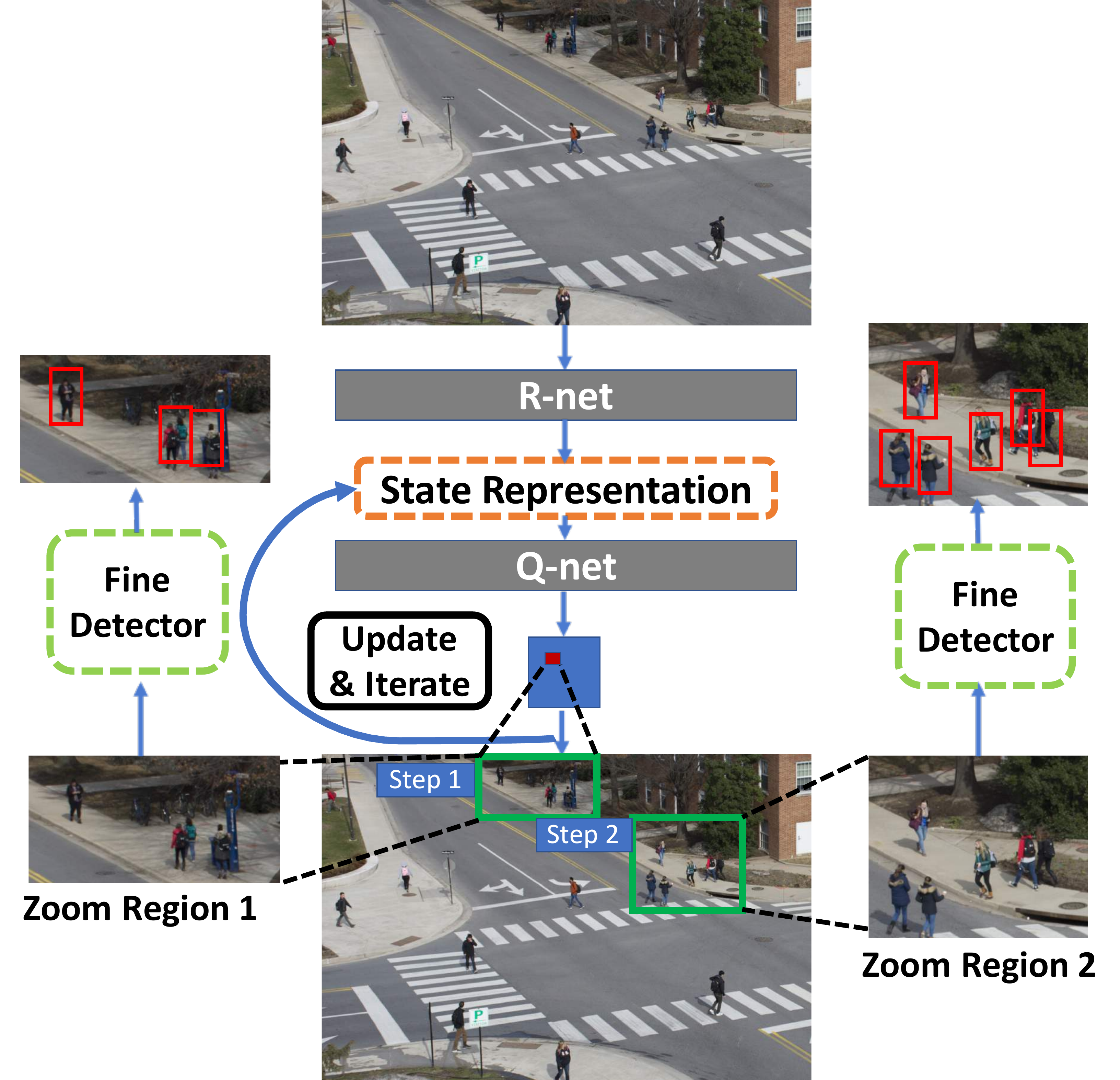}
\end{center}
   \caption{Illustration of our approach. The input is a down-sampled version of the image to which a coarse detector is applied. The R-net uses the initial coarse detection results to predict the utility of zooming in on a region to perform detection at higher resolution. The Q-net, then uses the computed accuracy gain map and a history of previous zooms to determine the next zoom that is most likely to improve detection with limited computational cost.}
\label{fig: idea}
\end{figure}
Most recent convolutional neural network (CNN) detectors are applied to images with relatively low resolution, \eg, VOC2007/2012 (about 500$\times$400)~\cite{pascal-voc-2007, pascal-voc-2012} and MS COCO (about 600$\times$400)~\cite{lin2014microsoft}. At such low resolutions, the computational cost of convolution is low. However, the resolution of everyday devices has quickly outpaced standard computer vision datasets. The camera of a 4K smartphone, for instance, has a resolution of 2,160$\times$3,840 pixels and a DSLR camera can reach 6,000$\times$4,000 pixels. Applying state-of-the-art CNN detectors directly to those high resolution images requires a large amount of processing time. Additionally, the convolution output maps are too large for the memory of current GPUs.

Prior works address some of these issues by simplifying the network architecture~\cite{PerforatedCNN, Nonlinear, DentonLeCun, Tucker, yu2017nisp} to speed up detection and reduce GPU memory consumption. However, these models are tailored to particular network structures and may not generalize well to new architectures. A more general direction is treating the detector as a black box that is judiciously applied to optimize accuracy and efficiency. For example, one could partition an image into sub-images that satisfy memory constraints and apply the CNN to each sub-image. However, this solution is still computationally burdensome. One could also speed up detection process and reduce memory requirements by running existing detectors on down-sampled images. However, the smallest objects may become too small to detect in the down-sampled images. Object proposal methods are the basis for most CNN detectors, restricting expensive analysis to regions that are likely to contain objects of interest~\cite{endres2010category, uijlings2013selective, zitnick2014edge, zhang2015bing++}. However, the number of object proposals needed to achieve good recall for small objects in large images is prohibitively high which leads to huge computational cost. 

Our approach is illustrated in Fig.~\ref{fig: idea}.
We speed up object detection by first performing coarse detection on a down-sampled version of the image and then sequentially selecting promising regions to be analyzed at a higher resolution. We employ reinforcement learning to model long-term reward in terms of detection accuracy and computational cost and dynamically select a sequence of regions to analyze at higher resolution. Our approach consists of two networks: a zoom-in accuracy gain regression network (R-net) learns correlations between coarse and fine detections and predicts the accuracy gain for zooming in on a region; a zoom-in Q function network (Q-net) learns to sequentially select the optimal zoom locations and scales by analyzing the output of the R-net and the history of previously analyzed regions.

Experiments demonstrate that, with a negligible drop in detection accuracy, our method reduces processed pixels by over $50\%$ and average detection time by $25\%$  on the Caltech Pedestrian Detection dataset~\cite{dollar2012pedestrian}, and reduces processed pixels by about $70\%$ and average detection time by over $50\%$ on a high resolution dataset collected from YFCC100M~\cite{kalkowski2015real} that has pedestrians of varied sizes. We also compare our method to recent single-shot detectors~\cite{redmon2016yolo9000, liu2016ssd} to show our advantage when handling large images.

\begin{figure*}[t]
\begin{center}
   \includegraphics[width=0.9\linewidth]{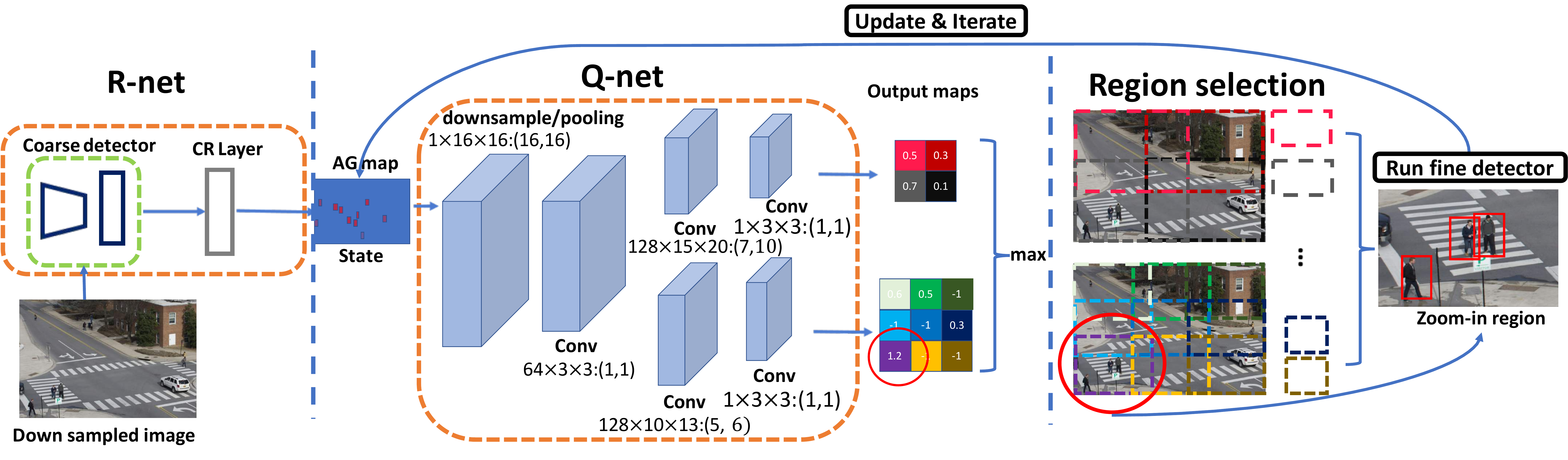}
\end{center}
   \caption{Given a down-sampled image as input, the R-net generates an initial accuracy gain (AG) map indicating the potential zoom-in accuracy gain of different regions (initial state). The Q-net is applied iteratively on the AG map to select regions. Once a region is selected, the AG map will be updated to reflect the history of actions. For the Q-net, two parallel pipelines are used, each of which outputs an action-reward map that corresponds to selecting zoom-in regions with a specific size. The value of the map indicates the likelihood that the action will increase accuracy at low cost. Action rewards from all maps are considered to select the optimal zoom-in region at each iteration. The notation 128$\times$15$\times$20:(7,10) means 128 convolution kernels with size 15$\times$20, and stride of 7/10 in height/width. Each grid cell in the output maps is given a unique color, and a bounding box of the same color is drawn on the image to denote the corresponding zoom region size and location.}
\label{fig: pipeline}
\end{figure*}
\section{Related work}
\label{sec: related works}
\textbf{CNN detectors.} One way to analyze high resolution images efficiently is to improve the underlying detector. Girshick~\cite{girshick2015fast} speeded up the region proposal based CNN~\cite{girshick14CVPR} by sharing convolutional features between proposals. Ren \etal proposed Faster R-CNN~\cite{ren2015faster}, a fully end-to-end pipeline that shares features between proposal generation and object detection, improving both accuracy and computational efficiency. Recently, single-shot detectors~\cite{liu2016ssd,redmon2016you,redmon2016yolo9000} have received much attention for real-time performance. These methods remove the proposal generation stage and formulate detection as a regression problem. Although these detectors performed well on PASCAL VOC~\cite{pascal-voc-2007, pascal-voc-2012} and MS COCO~\cite{lin2014microsoft} datasets, which generally contain large objects in images with relatively low resolution, they do not generalize as well on large images with objects of variable sizes. Also, their processing cost increases dramatically with image size due to the large number of convolution operations.

\textbf{Sequential search.} Another strategy to handle large image sizes is to avoid processing the entire image and instead investigate small regions sequentially. However, most existing works focus on mining informative regions to improve detection accuracy without considering computational cost. Lu~\etal~\cite{lu2015adaptive} improve localization by adaptively focusing on subregions likely to contain objects. Alexe \etal~\cite{alexe2012searching} sequentially investigated locations based on what has been seen to improve detection accuracy. However, the proposed approach introduces a large overhead leading to long detection time (about 5s per object class per image). Zhang \etal~\cite{zhang2015improving} improved the detection accuracy by penalizing the inaccurate location of the initial object proposals, which introduced more than $15\%$ overhead to detection time.

A sequential search process can also make use of contextual cues from sources, such as scene segmentation. Existing approaches have explored this idea for various object localization tasks~\cite{chen2016object, yu16objectcontextselection, mottaghi2014role}. Such cues can also be incorporated within our framework (\eg, as input to predicting the zoom in reward). However, we focus on using only coarse detections as a guide for sequential search and leave additional contextual information for future work. Other previous work~\cite{li2015convolutional} utilizes a coarse-to-fine strategy to speed up detection, but this work does not select promising regions sequentially.

\textbf{Reinforcement learning (RL).} RL a is popular mechanism for learning sequential search policies, as it allows models to consider the effect of a sequence of actions rather than individual ones. Ba \etal use RL to train a attention based model in~\cite{ba2014multiple} to sequentially select most relevant regions for object recognition and Jie~\etal~\cite{jie2016tree} select regions for localization in a top-down search fashion. However, these methods require a large number of selection steps and may lead to long running time. Caicedo~\etal~\cite{caicedo2015active} designed an active detection model for object localization, which utilizes Deep Q Networks (DQN)~\cite{mnih2015human} to learn a long-term reward function to transform an initial bounding box sequentially until it converges to an object. However, as reported in~\cite{caicedo2015active}, the box transformation takes about 1.5s detection time on a typical Pascal VOC image which is much slower than recent detectors~\cite{ren2015faster,liu2016ssd,redmon2016yolo9000}. In addition,~\cite{caicedo2015active} does not explicitly consider selection cost.
Although, RL implicitly forces the algorithm to take a minimum number of steps, we need to explicitly penalize cost since each step can yield a high cost. For example, if we do not penalize cost, the algorithm will tend to zoom in on the whole image. Existing works have proposed methods to apply RL in cost sensitive settings~\cite{he2012cost,karayev2014anytime}. We follow the approach of~\cite{he2012cost} and treat the reward function as a linear combination of accuracy and cost.

\section{Dynamic zoom-in network}
Our work employs a coarse-to-fine strategy, applying a coarse detector at low resolution and using the outputs of this detector to guide an in-depth search for objects at high resolution. The intuition is that, while the coarse detector will not be as accurate as the fine detector, it will identify image regions that need to be further analyzed, incurring the cost of high resolution detection only in promising regions. We make use of two major components: 1) a mechanism for learning the statistical relationship between the coarse and fine detectors, so that we can predict which regions need to be zoomed in given the coarse detector output; and 2) a mechanism for selecting a sequence of regions to analyze at high resolution, given the coarse detector output and the regions that have already been analyzed by the fine detector. Our pipeline is illustrated in Fig.~\ref{fig: pipeline}. We learn a strategy that models the long-term goal of maximizing the overall detection accuracy with limited cost.

\subsection{Problem formulation}
Our work is formulated as a Markov Decision Process (MDP)~\cite{Bel}. At each step, the system observes the current state, estimates potential cost-aware rewards of taking different actions and selects the action that has the maximum long-term cost-aware reward.

\textbf{Action.} Our algorithm sequentially analyzes regions with high zoom-in reward at high resolution. In this context, an \textit{action} corresponds to selecting a region to analyze at high resolution. Each action $a$ can be represented by a tuple $(x,y,w,h)$ where $(x, y)$ indicates the location, and $(w, h)$ specifies the size of the region. At each step, the algorithm scores a set of potential actions---a list of rectangular regions---in terms of the potential long-term reward of taking those actions.

\textbf{State.} The representation encodes two types of information: 1) the predicted accuracy gain of regions yet to be analyzed; and 2) the history of regions that have already been analyzed at high resolution (the same region should not be zoomed in multiple times). We design a zoom-in accuracy gain regression network (R-net) to learn an informative accuracy gain map (AG map) as the state representation from which the zoom-in Q function can be successfully learned. The AG map has the same width and height as the input image. The value of each pixel in the AG map is an estimate of how much the detection accuracy might be improved if that pixel in the input image were included by the zoom-in region. As a result, the AG map provides detection accuracy gain for selecting different actions. After an action is taken, values corresponding to the selected region in the AG map decrease accordingly, so the AG map can dynamically record action history.

\textbf{Cost-aware reward function.} The state representation encodes the predicted accuracy gain of zooming in on each image subregion. To maintain a high accuracy with limited computation, we define a cost-aware reward function for actions. Given state $s$ and action $a$, the cost-aware reward function scores each action (zoom region) by considering both cost increment and accuracy improvement as
\begin{equation}
\label{eq: r_function}
    R(s,a) = \sum_{k \ in \ a}{|g_k-p^l_k|-|g_k-p^h_k|}-\lambda\frac{b}{B}
\end{equation}
where $k \ in \ a$ means that proposal $k$ is included in the region selected by action $a$. $p^l_k$ and $p^h_k$ indicates coarse and fine detection scores, and $g_k$ is the corresponding ground-truth label. The variable $b$ represents the total number of pixels included in the selected region, and $B$ indicates the total number of pixels of the input image. The first term measures the detection accuracy improvement. The second term indicates the zoom-in cost. The trade-off between accuracy and computation is controlled by the parameter $\lambda$. During training, the Q-net uses this reward function to calculate the immediate rewards of taking actions and learns a long-term reward function by Q learning~\cite{watkins1992q}. 
\subsection{Zoom-in accuracy gain regression network}
\label{sec: r-net}
The zoom-in accuracy gain regression network (R-net) predicts the accuracy gain of zooming in on a particular region based on the coarse detection results. The R-net is trained on pairs of coarse and fine detections so that it can observe how they correlate with each other to learn a suitable accuracy gain. 

Toward this end, we apply two pre-trained detectors to a set of training images and obtain two sets of image detection results: low-resolution detections $\{(\mathbf{d}^l_i, p^l_i, \mathbf{f}^l_i)\}$ in the down-sampled image and high-resolution detections $\{(\mathbf{d}^h_j, p^h_j)\}$ in the high resolution version of each image, where $\mathbf{d}$ is the detection bounding box, $p$ is the probability of being the target object and $\mathbf{f}$ indicates a feature vector of the corresponding detection. We use the superscripts $h$ and $l$ to indicate the high resolution and low resolution (down-sampled) images. For the model to learn whether or not a high resolution detection improves the overall results, given a set of coarse detections at training time, we introduce a \emph{match layer} which associates detections produced by the two detectors. In this layer, we pair the coarse and fine detection proposals and generate a set of correspondences between them. The object proposals $i$ in the down-sampled image and $j$ in the high-resolution image are defined as corresponding to each other if we find a $j$ with sufficiently large intersection over union $\text{IoU}(d^l_i, d^h_j)$ with $i$ ($\text{IoU}>$0.5).

Given a set of correspondences, $\{(\mathbf{d}^l_k, p^l_k, p^h_k, \mathbf{f}^l_k)\}$, we estimate the zoom-in accuracy gain of a coarse detection. A detector can handle only objects within a range of sizes, so applying the detector to the high-resolution image does not always produce the best accuracy. For example, larger objects might be detected with higher accuracy at lower resolution if the detector was trained on mostly smaller objects. So, we measure which detection (coarse or fine) is closer to groundtruth using the metric $|g_k-p^l_k|-|g_k-p^h_k|$ where $g_k \in \{0, 1\}$ indicates the groundtruth label. When the high resolution score $p^h_k$ is closer to the groundtruth than the low resolution score $p^l_k$, the function indicates that this proposal is worth zooming in on. Otherwise, applying a detector on the down-sampled image is likely to yield a higher accuracy, so we should avoid zooming in on this proposal. We use a Correlation Regression (CR) layer to estimate the zoom-in accuracy gain of proposal $k$ such that
\begin{equation}
\label{eq: regression}
\underset{\mathbf{W}}{\min}(|g_k-p^l_k|-|g_k-p^h_k|-\Phi(\mathbf{W}, f_k^l))^2~,
\end{equation}
where $\Phi$ represents the regression function and $\mathbf{W}$ indicates the parameters.
The output of this layer is the estimated accuracy gain. The CR layer contains two fully connected layers where the first layer has 4,096 units and the second one has only one output unit.

The AG map can be generated given the learned accuracy gain of each proposal. We assume that each pixel inside a proposal bounding box has equal contribution to its accuracy gain. Consequently, the AG map is generated as
\begin{equation}
\label{eq: ra}
  AG(x,y)=\begin{cases}\alpha\frac{\Phi(\mathbf{\widehat{W}}, f_k^l)}{b_k}
 & \textit{if } (x,y) \ in \ \mathbf{d}^l_k,\\
 0
 & \textit{otherwise},
\end{cases}
\end{equation}
where $(x,y) \ in \ \mathbf{d}^l_k$ means point $(x,y)$ is inside the bounding box $\mathbf{d}^l_k$ and $b_k$ denotes the number of pixels included in $\mathbf{d}^l_k$. $\alpha$ is a constant number. $\mathbf{\widehat{W}}$ denotes the estimated parameters of the CR layer.
The AG map is used as the state representation and it naturally contains the information of coarse detections' qualities. After zooming in and performing detection on a region, all the values inside the region are set 0 to prevent future zooming on the same region.
\subsection{Zoom-in Q function learning network}
\label{sec: q-net}

The R-net provides information about which image region is likely to be the most informative if it is inspected next. Since the R-net is embedded within a sequential process, we use reinforcement learning to train a second network, the Q-net, to learn a long-term zoom-in reward function. At each step, the system takes an action by considering both immediate (Eq.~\ref{eq: r_function}) and future rewards. We formulate our problem in a Q learning framework, which approximates the long-term reward function for actions by learning a Q function. Based on the Bellman equation~\cite{bellman1956dynamic}, the optimal Q function, $Q^*(s,a)$, obeys an important identity: given the current state, the optimal reward of taking an action equals the combination of its immediate reward and a discounted optimal reward at the next state triggered by this action~\eqref{eq: bellman} 
\begin{equation}
Q^*(s,a)=\mathbb{E}_{s'}[R(s,a)+\gamma\max_{a'} Q{^*}({s'},{a'})|s,a]\label{eq: bellman}
\end{equation}
where $s$ is the state and $a$ is an action.
Following~\cite{mnih2015human}, we learn the Q function for candidate actions by minimizing the loss function at the $i$-th iteration, \ie,
\begin{align}
\label{eq: q_loss}
L_i=&(R(s,a)+\gamma \underset{a'}{\max} Q({s'},{a'};\theta^-_i) - Q({s},{a};\theta_i))^2
\end{align}
where $\theta_i$ and $\theta^-_i$ are parameters of the Q network and those needed to calculate future reward at iteration $i$, respectively. 

Eq.~\ref{eq: q_loss} implies that the optimal long-term reward can be learned iteratively if the immediate reward $R(s,a)$ is provided for a state-action pair. Since $R(s, a)$ is a cost-aware reward, the Q-net learns a long-term cost-aware reward function for the action set.

In practice, $\theta^-_i = \theta_{i-C}$ where $C$ is a constant parameter. $\gamma$ is future reward discount factor. We choose $C=10$ and $\gamma = 0.5$ empirically in our experiments. We also adopt the $\epsilon$-greedy policy~\cite{sutton1998reinforcement} at training to balance between exploration and exploitation. The $\epsilon$ setting is the same as in~\cite{caicedo2015active}.

The structure of our Q-net is shown in Fig.~\ref{fig: pipeline}. The input is the AG map and each pixel in the map measures the predicted accuracy gain if the pixel at that location in the input image is included in the zoom region. The output is a set of maps and each value of a map measures the long-term reward of taking the corresponding action (selecting a zoom region at a location with a specified size). To allow the Q-net to choose zoomed-in regions with different sizes, we use multiple pipelines, each of which outputs a map corresponding to zoomed-in regions of a specific size. These pipelines share the same features extracted from the state representation. In the training phase, actions from all maps are concatenated to produce a unified action set and trained end-to-end together by minimizing the loss function in Eq.~\ref{eq: q_loss} so that all actions values compete with each other.

After zooming in on a selected region, we get both coarse and fine detections on the region. We just replace the coarse detections with fine ones in each zoom-in region.

\textbf{Window selection refinement.} The output of the Q-net can be directly used as a zoom-in window. However, because candidate zoom windows are sparsely sampled, the window can be adjusted slightly to increase the expected reward. The Refine module takes the Q-net output as a coarse selection and locally moves the window towards a better location, as measured by the accuracy gain map by 
\begin{equation}
\label{eq: refine}
\hat{a} = \arg \underset{a \in A}{\max}\sum_{(x,y) \ in \ a} AG(x,y)
\end{equation}
where $\hat{a}$ selects the refined window and $A =$ ${(x_q\pm \mu_x , y_q\pm\mu_y, w, h) }$ corresponds to the local refinement area controlled by parameter $\mu$, where ${(x_q , y_q, w, h) }$ indicates the output window of Q-net. We show a qualitative example of refinement in Fig.~\ref{fig: refine}.

\section{Experiments}
\label{sec: experiments}
We perform experiments on the Caltech Pedestrian Detection dataset (CPD)~\cite{dollar2012pedestrian} and a Web Pedestrian dataset (WP) collected from YFCC100M~\cite{kalkowski2015real}. Datasets like Pascal VOC~\cite{pascal-voc-2007} and MS COCO~\cite{lin2014microsoft} are not chosen to validate our method, because they are not close to our scenario. In \cite{pascal-voc-2007} and \cite{lin2014microsoft}, there are generally very few objects per image and most objects are large, which leads to 1) close-to-zero rewards for regions, since large objects are likely to maintain high detection accuracy after reasonable down sampling; and, 2) large zoom-in windows in order to enclose large objects. Low region rewards discourage the window selection process and large zoom-in windows produce high cost, which make our method invalid.

\textbf{Caltech Pedestrian Detection (CPD).} 
There are different settings according to different annotation types, \ie \textit{Overall}, \textit{Near scale}, \textit{Medium scale}, \textit{No occlusion}, \textit{Partial occlusion} and \textit{Reasonable}~\cite{dollar2012pedestrian}. Similar to the \textit{Reasonable} setting, we only train and test on pedestrians at least 50 pixels tall. We sparsely sample images (every 30 frames) from the training set. There are 4,321 images in the training set and 4,088 images in the test set. We rescale the images to 600 pixels on the shorter side to form the high resolution version of image during both training and testing. All of our model components are trained on this training set.

\textbf{Web Pedestrian (WP) dataset.} The image resolution in the CPD dataset is low (640$\times$480). To better demonstrate our approach, we collect 100 test images with much higher resolution from the YFCC100M~\cite{kalkowski2015real} dataset. The images are collected by searching for keywords "Pedestrian", "Campus" and "Plaza". An example is shown in Fig.~\ref{fig: gsvsq} where pedestrians have varied sizes and are densely distributed in the images. For this dataset, we annotate all the pedestrians with at least 16-pixel width and less than $50\%$ occlusion. Images are rescaled to 2,000 pixels on the longer side to fit for our GPU memory.

\subsection{Baseline methods}
We compare to the following baseline algorithms:

\textbf{\textit{Fine-detection-all}.} This baseline directly applies the fine detector to the high resolution version of image. This method leads to high detection accuracy with high computational cost. All of the other approaches seek to maintain this detection accuracy with less computation.

\textbf{\textit{Coarse-detection-all}.} This baseline applies the coarse detector on down-sampled images with no zooming.

\textbf{\textit{GS+Rnet}.} Given the initial state representation generated by the R-net, we use a greedy search strategy (GS) to densely search for the best window every time based on the current state without considering the long-term reward. 

\textbf{\textit{ER+Qnet}}. The entropy  of the detector output (object vs no object) is another way to measure the quality of a coarse detection. \cite{almahairi2015dynamic} used entropy to measure the quality of a region for a classification task. Higher entropy implies lower quality of a coarse detection. So, if we ignore the correlation between fine and coarse detections, the accuracy gain of a region can also be computed as 
\begin{equation}
\label{eq: entropy}
-p^l_ilog(p^l_i)-(1-p^l_i)log(1-p^l_i)
\end{equation}
where $p^l$ indicates the score of the coarse detection.
For fair comparison, we fix all parameters of the pipeline except replacing the R-net output of a proposal with its entropy.
\begin{figure}[t]
\centering
   \includegraphics[width=0.8\linewidth]{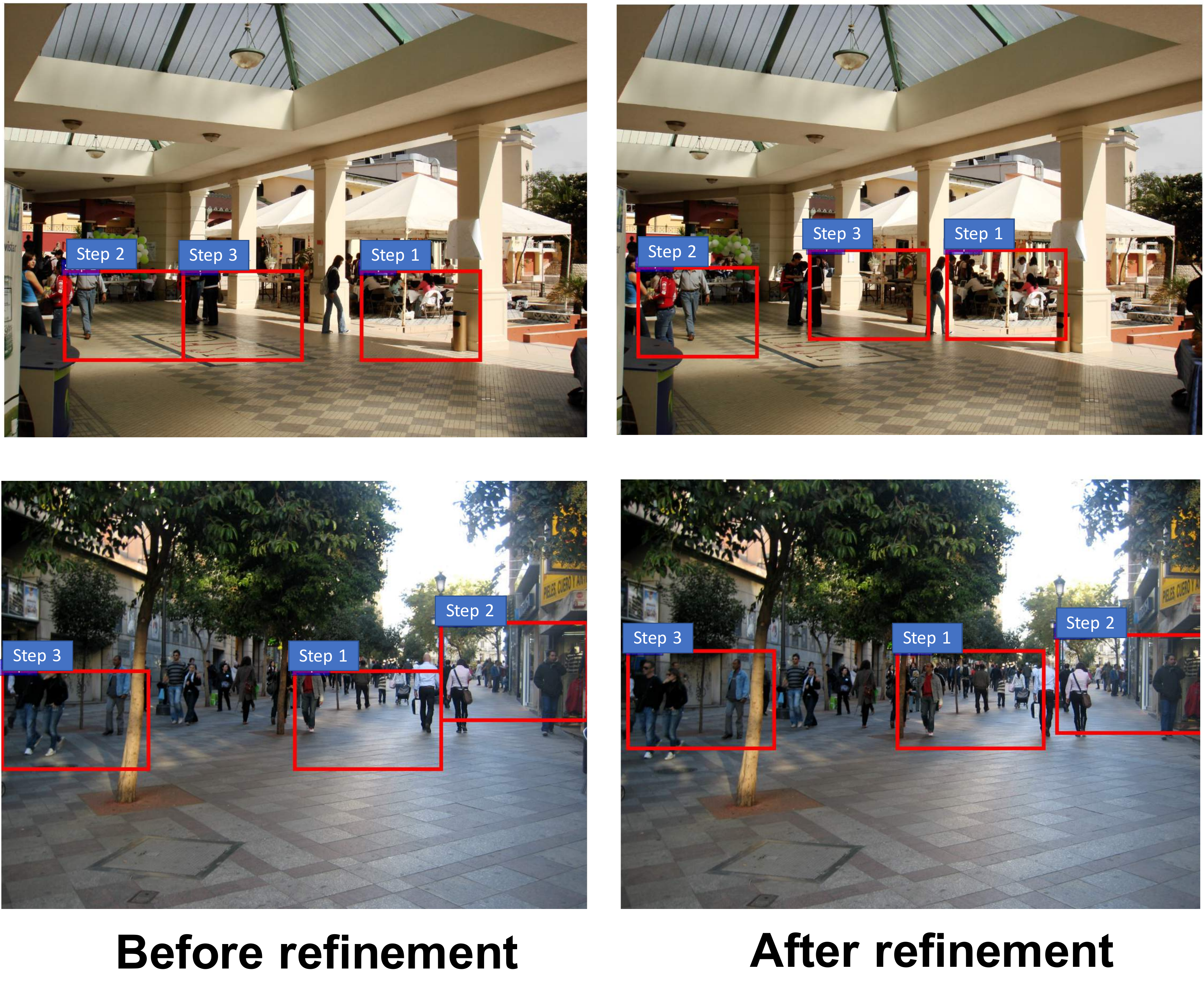}
   \caption{Effect of region refinement. Red boxes indicate zoom regions and the step number denotes the order that the zoom windows were selected. Before refinement, windows are likely to cut people in half due to the sampling grid, leading to a bad detection performance. Refinement locally adjusts the location of a window and produces better results.}
\label{fig: refine}
\end{figure}

\textbf{SSD and YOLOv2.} We also compare our method with off-the-shelf SSD~\cite{liu2016ssd} and YOLOv2~\cite{redmon2016yolo9000} trained on CPD, to show the advantage of our method on large images.
\subsection{Variants of our framework}
We use \textit{Qnet-CNN} to represent the Q-net developed using a fully convolutional network (see Fig.~\ref{fig: pipeline}). To analyze the contributions of different components to the performance gain, we evaluate three variants of our framework: Qnet*, Qnet-FC and Rnet*.

\textbf{\textit{Qnet*}.} This method uses a Q-net with refinement to locally adjust the zoom-in window selected by Q-net.

\textbf{\textit{Qnet-FC}.} Following \cite{caicedo2015active}, we develop this variant with two fully connected (FC) layers for Q-net. For \textit{Qnet-FC}, the state representation is resized to a vector of length $1,200$ as the input. The first layer has 128 units and the second layer has 34 units (9+25). Each output unit represents a sampled window on an image. We uniformly sample 25 windows of size $320 \times 240$ and 9 windows of size $214 \times 160$ on the CPD dataset. Since the output number of \textit{Qnet-FC} can not be changed, windows sizes are proportionally increased when \textit{Qnet-FC} is applied to WP dataset.

\textbf{\textit{Rnet*}.} This is an R-net learned using a reward function that does not explicitly encode cost ($\lambda=0$ in Eq.~\ref{eq: r_function}).

\subsection{Evaluation metric}
We use three metrics when comparing to the \textit{Fine-detection-all} strategy: AP percentage ($A_{perc}$), processed pixel numbers percentage ($P_{perc}$), and average detection time percentage ($T_{perc}$). $A_{perc}$ quantifies the percentage of AP we obtain compared to the \textit{Fine-detection-all} strategy. $P_{perc}$ and $T_{perc}$ indicate the computational cost as a percentage of the \textit{Fine-detection-all baseline} strategy.

\subsection{Implementation details}
We downsample the high resolution image by a factor of $2$ to form a down-sampled image for all of our experiments and only handle zoom-in regions at the high resolution.

For the Q-net, we spatially sample zoom-in candidate regions with two different window sizes ($320 \times 240$ and $214 \times 160$) in a sliding window manner. For windows of size $W\times H$, we uniformly sample windows with horizontal stride $S_x=W/2$ and vertical stride $S_y=H/2$ pixels. For the refinement, we set $(\mu_x,\mu_y)=0.5(S_x, S_y)$. The Q-net stops taking actions when the sum over all the values of the AG map is smaller than $0.1$.

We use Faster R-CNN as our detector due to the success of R-CNN in many computer vision applications~\cite{gao2017cwsl,yu2017_vrd_knowledge_distillation,li2017_3dscenefromtext,Yu2018_ReMotENet,DBLP:journals/corr/HeGDG17,DBLP:journals/corr/BellZBG15}. Two  
Faster R-CNNs are trained on the CPD training set at the fine and coarse resolutions and used as black-box coarse and fine detectors afterwards. YOLOv2 
and SSD 
are trained on the same training set with default parameter settings in the official codes released by the authors. All experiments are conducted using a K-80 GPU.

\subsection{Qualitative results}\label{qual}
The qualitative comparisons, which show the effect of refinement on the selected zoom-in regions, are shown in Fig. \ref{fig: refine}. We observe that refinement significantly reduces the cases in which pedestrians only partly occur in the selected windows. Due to the sparse window sampling of Q-net, optimal regions might not be covered by any window candidate, especially when the window size is relatively small compared to the image size.

\begin{figure}[t]
\centering
   \includegraphics[width=0.8\linewidth]{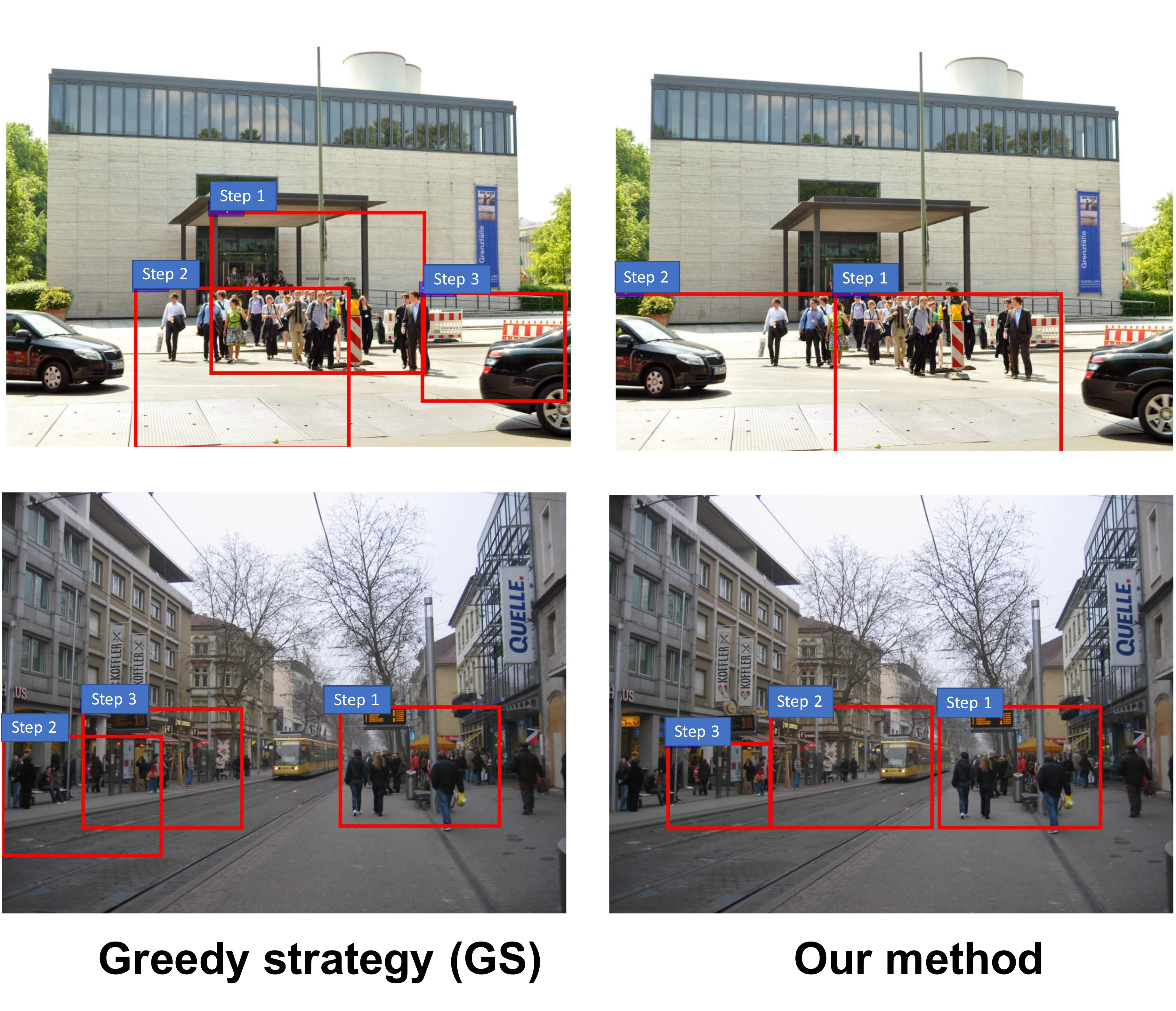}
   \caption{Qualitative comparison between using the Q-net* and a greedy strategy (GS) that selects the region with highest predicted accuracy gain at each step. Red bounding boxes indicate zoom-in windows and step number denotes the order of windows selection. The Q-net selects regions that appear sub-optimal in the near term but better zoom sequences in the long term, which leads to fewer steps as shown in the first row.}
\label{fig: gsvsq}
\end{figure}

We show a comparison between our method (\textit{Q-net*-CNN+Rnet}) and the greedy strategy (\textit{GS+Rnet}) in Fig.~\ref{fig: gsvsq}. \textit{GS} tends to select duplicate zooms on the same portion of the image. While the Q-net might select a sub-optimal window in the near term, it leads to better overall performance in the long term. As shown in the first example of Fig.~\ref{fig: gsvsq}, this helps Q-net terminate with fewer zooms.

\begin{figure}[t]
\centering
   \includegraphics[width=1.0\linewidth]{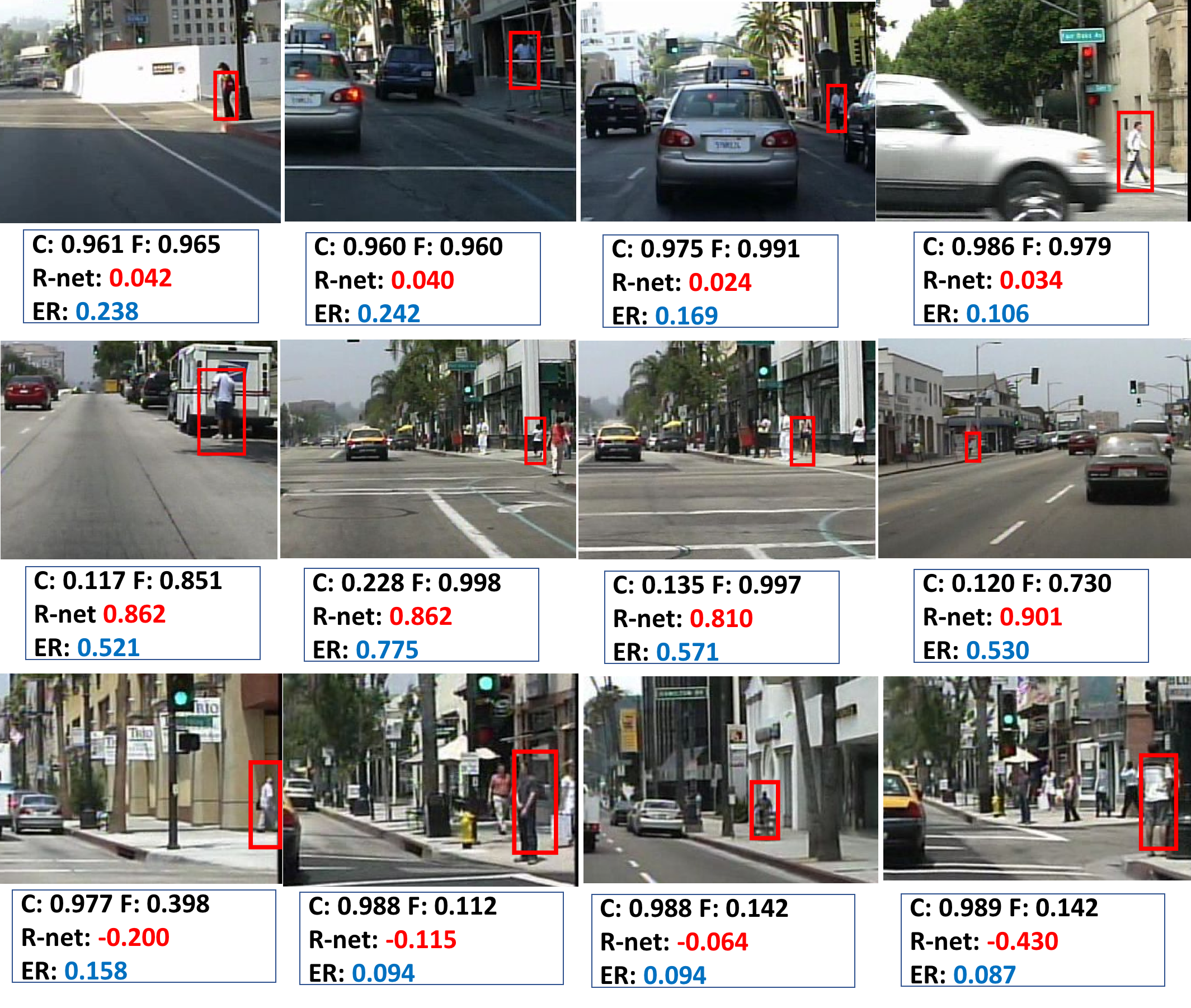}
   \caption{Qualitative comparison of R-net and ER on the Caltech Pedestrians test set. The first row of numbers indicate probability of the red box being a pedestrian. C denotes coarse detection and F indicates fine detection. Red font denotes the accuracy gain of R-net and blue is for ER. Positive and negative values are normalized to [0, 1] and [-1, 0). Compared to ER, R-net gives lower positive scores (row $\#$1)/ negative scores (row $\#$3) for regions that coarse detections are good enough/ better than fine detections and it produces higher scores for regions (row $\#$2) where fine detections are much better than coarse ones.} 
\label{fig: rnet_qualitative}
\end{figure}

Fig.~\ref{fig: rnet_qualitative} shows a qualitative comparison of R-net and \textit{ER}. The examples in the first row are detections that do not need to be zoomed in on, since the coarse detections are good enough. R-net produces much lower accuracy gains for these regions. On the other hand, R-net outputs much higher gains in the second row which includes regions needing analysis at higher resolution. The third row contains examples which get worse results at higher resolution. As we mentioned before, entropy cannot determine if zooming in will help, while R-net produces negative gains for these cases and avoids zooming in on these regions.

\subsection{Quantitative evaluation}

Table \ref{tab: coarse_fine} shows the average precision (AP) and average detection time per image for \textit{Fine-detection-all} and \textit{Coarse-detection-all} strategies on CPD and WP datasets. The coarse baseline maintains only about $65\%$ and $71\%$ AP on CPD and WP, respectively, suggesting that the naive downsamping method significantly decreases detection accuracy.

\begin{table}[]
\small
\centering
\begin{tabular}{|c|c|c|c|c|}
\hline
            Dataset& $\text{AP}_f$ & {$\text{AP}_c$} & $\text{DT}_f$(ms) & $\text{DT}_c$(ms) \\ \hline
CPD & 0.493 & 0.322 & 304  & 123      \\ \hline
WP & 0.407 & 0.289 &1375 &427            \\ \hline
\end{tabular}
\caption{\textit{Coarse-detection-all}(with subscript $c$) v.s. \textit{Fine-detection-all} (with subscript $f$) on CPD and WP datasets. DT indicates average detection time per image.}
\label{tab: coarse_fine}
\end{table}
\begin{figure}[t]
\centering
   \includegraphics[width=1.0\linewidth]{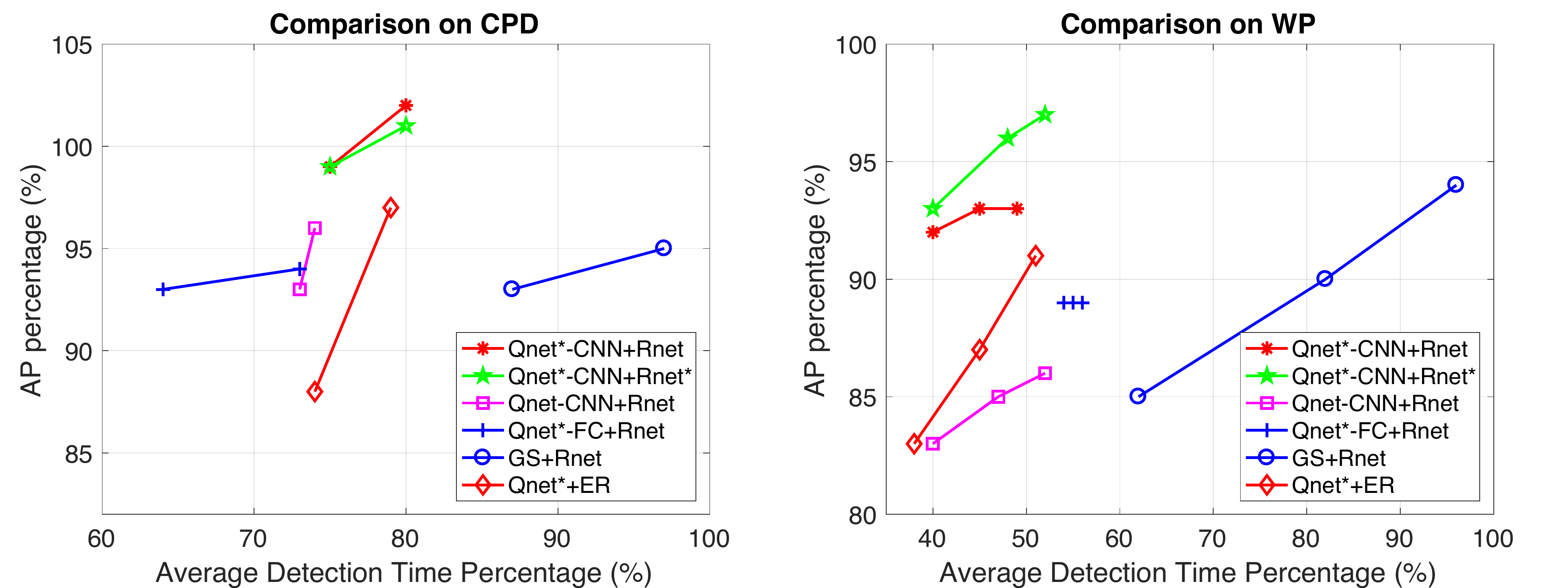}
   \caption{Detection time and accuracy comparison on the CPD/WP dataset after zooming in on two/three regions.}
\label{fig: time_combine}
\end{figure}
\begin{table*}[]
\small
\centering
\begin{tabular}{|l|c|c|l|c|c|c|c|c|}
\hline
\multirow{2}{*}{} &  & \multicolumn{3}{c|}{\bf Baselines} & \multicolumn{4}{c|}{\bf Variants under our framework} \\ \cline{2-9} 
 & $P_{perc}$ & \multicolumn{2}{c|}{GS+Rnet} & Qnet*-CNN+ER & Qnet*-CNN+Rnet & Qnet*-CNN+Rnet* & Qnet*-FC+Rnet & Qnet-CNN+Rnet \\ \hline
\multirow{3}{*}{\bf CPD} & $\leq$40\% & \multicolumn{2}{c|}{65\%(40\%)} & 88\%(74\%) & \textbf{99\%}(75\%) & 65\%(40\%) & 93\%(64\%) & 65\%(40\%) \\ \cline{2-9} 
 & $\leq$45\% & \multicolumn{2}{c|}{93\%(87\%)} & 97\%(79\%) & \textbf{102\%}(80\%) & 101\%(80\%) & 94\%(73\%) & 96\%(73\%) \\ \cline{2-9} 
 & $\leq$50\% & \multicolumn{2}{c|}{95\%(97\%)} & 97\%(79\%) & \textbf{102\%}(80\%) & 101\%(80\%) & 94\%(73\%) & 96\%(73\%) \\ \hline
\multirow{3}{*}{\bf WP} & $\leq$30\% & \multicolumn{2}{c|}{85\%(62\%)} & 83\%(38\%) & 92\%(40\%) & \textbf{93\%}(40\%) & 71\%(31\%) & 83\%(40\%) \\ \cline{2-9} 
 & $\leq$35\% & \multicolumn{2}{c|}{90\%(82\%)} & 91\%(51\%) & 93\%(45\%) & \textbf{96\%}(48\%) & 71\%(31\%) & 85\%(47\%) \\ \cline{2-9} 
 & $\leq$40\% & \multicolumn{2}{c|}{94\%(96\%)} & 91\%(51\%) & 93\%(49\%) & \textbf{97\%}(52\%) & 89\%(54\%) & 86\%(52\%) \\ \hline
\end{tabular}
\caption{Detection accuracy comparisons in terms of $A_{perc}$ on the CPD and WP datasets under a fixed range of processed pixel percentage ($P_{perc}$). Bold font indicates the best result. Numbers are display as $A_{perc}(T_{perc})$- $T_{perc}$ is included in the parentheses for the reference of running time. Note that $25\%$ $P_{perc}$ overhead is incurred simply by analyzing the down-sampled image (this overhead is included in the table) and percentages are relative to \textit{Fine-detection-all} baseline (an $A_{perc}$ of $80\%$ means that an approach reached $80\%$ of the AP reached by the baseline). 
}
\label{tab: combine_PPP}
\end{table*}
Comparative results on the CPD and WP dataset are shown in Table~\ref{tab: combine_PPP}. \textit{Q-net*-CNN + R-net} reduces processed pixels by over $50\%$ with comparable (or even better) detection accuracy than the \textit{Fine-detection-all} strategy and improves detection accuracy of \textit{Coarse-detection-all} by about $35\%$ on the CPD dataset. On the WP dataset, the best variant (\textit{Q-net*-CNN + R-net*)}) reduces processed pixels by over $60\%$ while maintaining $97\%$ detection accuracy of \textit{Fine-detection-all}.
\begin{table}[]
\centering
\small
\begin{tabular}{|c|c|c|c|c|}
\hline
\multirow{2}{*}{} & \multicolumn{2}{c|}{\bf CPD} & \multicolumn{2}{c|}{\bf WP} \\ \cline{2-5} 
 & AP & DT(ms) & AP & DT(ms) \\ \hline
SSD500~\cite{liu2016ssd} & 0.405 & 128 & 0.255 & 570 \\ \hline
SSD300~\cite{liu2016ssd} & 0.400 & 74 & 0.264 & 530 \\ \hline
YOLOv2~\cite{redmon2016yolo9000} & 0.398 & 70 & 0.261 & 790 \\ \hline
Our method & \textbf{0.503} & 243 & \textbf{0.379} & 619 \\ \hline
\end{tabular}
\caption{Comparison between Qnet*-CNN+Rnet and single-shot detectors trained on CPD. DT indicates average detection time per image. Bold font indicates the best result.}
\label{tab: single-shot}
\end{table}
Table~\ref{tab: combine_PPP} shows that variants of our framework outperform \textit{GS+Rnet} and \textit{Qnet+ER} in most cases which suggests that \textit{Qnet} and \textit{Rnet} are better than \textit{GS} and \textit{ER}. Q-net is better than \textit{GS} since the greedy strategy considers individual actions separately, while Q-net utilizes a RL framework to maximize the long term reward.

\textit{Qnet*-CNN+Rnet} always produces better detection accuracy than \textit{Qnet*-CNN+ER} under the same cost budget, which demonstrates that learning the accuracy gain using an R-net is preferable to using entropy, a hand-crafted measure. This could be due to two reasons: 1) entropy measures only the confidence of the coarse detector, while our R-net estimates the correlation with the high-resolution detector based on confidence and appearance; 2) according to the regression target function in Eq.~\ref{eq: regression}, our R-net also measures whether the zoom-in process will improve detection accuracy. This avoids wasting resources on regions that cannot be improved (or might even be degraded) by fine detections. 

We observe from Fig.~\ref{fig: time_combine} that our approach (\textit{Qnet*-CNN+Rnet} and \textit{Qnet*-CNN+Rnet*}) reduces detection time by $50\%$ while maintaining a high accuracy on the WP dataset. On the CPD dataset, they can reduce detection time by $25\%$ without a significant drop of accuracy. Detection time cannot be reduced as much as on the WP dataset, since CPD images are relatively small; however, it is notable that our approach helps even in this case.

Table~\ref{tab: single-shot} shows accuracy/cost comparisons between YOLO/SSD and our method. Experiments suggest the following conclusions: 1) although fast, these single-shot detectors achieve much lower AP on images with objects occurring over a large range of scales; 2) as image size increases, YOLO/SSD processing time increases dramatically, while, our method achieves much higher accuracy with comparable detection time; 3) SSD consumes much more GPU memory than other detectors on large images due to the heavy convolution operations. We have to resize images of WP to $800\times800$ to fit within GPU memory. Note that it is possible to improve the results of YOLO/SSD by pruning the networks or training with more data, but that is not within the scope of this paper.
\subsection{Ablation analysis}
{\textbf{Improvement by refinement (\textit{Qnet*-CNN+Rnet} vs. \textit{Qnet-CNN+Rnet}).}} In Table~\ref{tab: combine_PPP}, we find that region refinement significantly improves detection accuracy under fixed cost ranges, especially on the WP. Refinement is more useful when zoom-in window size is relatively small compared with image size due to the sparse window sampling of Q-net. Fig.~\ref{fig: refine} qualitatively shows the effect of refinement.

{\textbf{Improvement by CNN (\textit{Qnet*-CNN+Rnet} vs. \textit{Qnet*-FC+Rnet}).}} 
FC has two obvious drawbacks in our setting. First, it has a fixed number of inputs and outputs which makes it hard to handle images with different sizes. Second, it is spatially dependent. Images from the CPD dataset consist of driving views which have strong spatial priors, \ie, most pedestrians are on the sides of the street and the horizon is roughly in the same place. \textit{Qnet-FC} takes advantage of these spatial priors, so it works better on this dataset. However, when it is applied to the WP dataset, its performance drops significantly compared to other methods, since the learned spatial priors now distract the detector.

{\textbf{Improvement by the cost term (\textit{Qnet*-CNN+Rnet} vs. \textit{Qnet*-CNN+Rnet*}).}} \textit{Qnet*-CNN+Rnet} outperforms \textit{Qnet*-CNN+R-net*} on CPD, especially when $P_{perc}$ is low ($40\%$). Without explicit cost penalization, the algorithm often selects the largest zoom regions, a poor strategy when there is a low pixel budget. However, since the window sizes are relatively small compared to the image size of the WP dataset,  \textit{Qnet*-CNN+Rnet*} does not suffer much from this limitation. On the contrary, it benefits from zooming in on relatively bigger regions. Consequently, it outperforms other variants. Nevertheless, \textit{Qnet*-CNN+Rnet} has comparable detection accuracy and can generalize better on scenarios where window sizes are comparable with image size.

\section{Conclusion}
{We propose a dynamic zoom-in network to speed up object detection in large images without manipulating the underlying detector's structure. Images are first downsampled and processed by the R-net to predict the accuracy gain of zooming in on a region. Then, the Q-net sequentially selects regions with high zoom-in reward to conduct fine detection. The experiments show that our method is effective on both Caltech Pedestrian Detection dataset and a high resolution pedestrian dataset.}
\\ 
\textbf{Acknowledgement.} This work was partially supported by the Defense Advanced Research Projects Agency (DARPA) under contract no.~W911NF-16-C-0003. The views and conclusions contained in this paper are those of the authors and should not be interpreted as representing the official policies, either expressly or implied, of DARPA or the U.S. Government.
{\small
\bibliographystyle{ieee}
\bibliography{egbib}
}

\end{document}